\pdfoutput=1

\documentclass[11pt]{article}

\usepackage[]{ACL2023}

\usepackage{times}
\usepackage{latexsym}
\usepackage{booktabs}
\usepackage{makecell}
\usepackage{pifont}
\usepackage{graphicx}
\usepackage{xspace}
\usepackage{amssymb}
\usepackage{amsfonts}
\usepackage{amsmath} 
\usepackage{amsthm} 
\usepackage{multirow}
\usepackage{colortbl}
\usepackage{appendix}
\usepackage{hyperref}
\usepackage{marvosym}

\makeatletter
\def\@fnsymbol#1{}
\makeatother

\usepackage[T1]{fontenc}

\usepackage[utf8]{inputenc}

\usepackage{microtype}

\usepackage{inconsolata}

\newcommand{\paratitle}[1]{\vspace{1.5ex}\noindent\textbf{#1}}
\newcommand{\ie}{\emph{i.e.,}\xspace}

\newcommand{\eg}{\emph{e.g.,}\xspace}

\newcommand{\ignore}[1]{}

\title{Zero-shot Visual Question Answering with Language Model Feedback}

\author{
\setcounter{footnote}{1}
	Yifan Du\textsuperscript{\rm{1},\rm{4}},
    Junyi Li\textsuperscript{\rm{1},\rm{3}},
	Tianyi Tang\textsuperscript{\rm{1}},
	\textbf{Wayne Xin Zhao}\textsuperscript{\rm{1},\rm{4} \Letter}\thanks{\Letter\ Corresponding author.} \and
	\textbf{Ji-Rong Wen}\textsuperscript{\rm{1}, \rm{2},\rm{4}} \\
        \textsuperscript{1}Gaoling School of Artificial Intelligence, Renmin University of China \\
	\textsuperscript{2}School of Information, Renmin University of China \\
        \textsuperscript{3}DIRO, Université de Montréal \\
	\textsuperscript{4}Beijing Key Laboratory of Big Data Management and Analysis Methods\\
	 \texttt{\{yifandu1999,                           
              batmanfly\}@gmail.com}\\ \texttt{lijunyi@ruc.edu.cn},
            \texttt{steventianyitang@outlook.com}
}

\begin{document}
\maketitle
\begin{abstract}
In this paper, we propose a novel language model guided captioning approach, \textbf{\textsc{Lamoc}},  for knowledge-based visual question answering~(VQA). Our approach employs the generated captions by a captioning model as the context of an answer prediction model, which is a Pre-trained Language model~(PLM). 
As the major contribution, we  leverage the guidance and feedback of the prediction model to improve the capability of the captioning model. In this way, the captioning model can become aware of the task goal and information need from the PLM. 
To develop our approach, we design two specific training stages, where the first stage adapts the captioning model to the prediction model (selecting more suitable caption propositions for training) and the second stage tunes the  captioning model according to the task goal (learning from feedback of the PLM). 
Extensive experiments demonstrate the effectiveness of the proposed approach on the knowledge-based VQA task. Specifically, on the challenging A-OKVQA dataset, \textsc{Lamoc} outperforms several competitive zero-shot methods and even achieves comparable results to a fine-tuned VLP model. Our code is publicly available at~\url{https://github.com/RUCAIBox/LAMOC}.

\end{abstract}
\section{Introduction}  

Recently, pre-trained language models~(PLMs)~\citep{devlin2019bert, brown2020language}, especially large language models~\citep{zhao2023survey} have demonstrated excellent capabilities in solving  tasks that require background  knowledge or complex reasoning, such as commonsense reasoning~
\citep{sap2019atomic, rajani2019explain} 
and logical reasoning~\citep{wei2022chain, kojima2022large}. 
\begin{figure}[ht]
    \centering
    \includegraphics[width=0.5\textwidth]{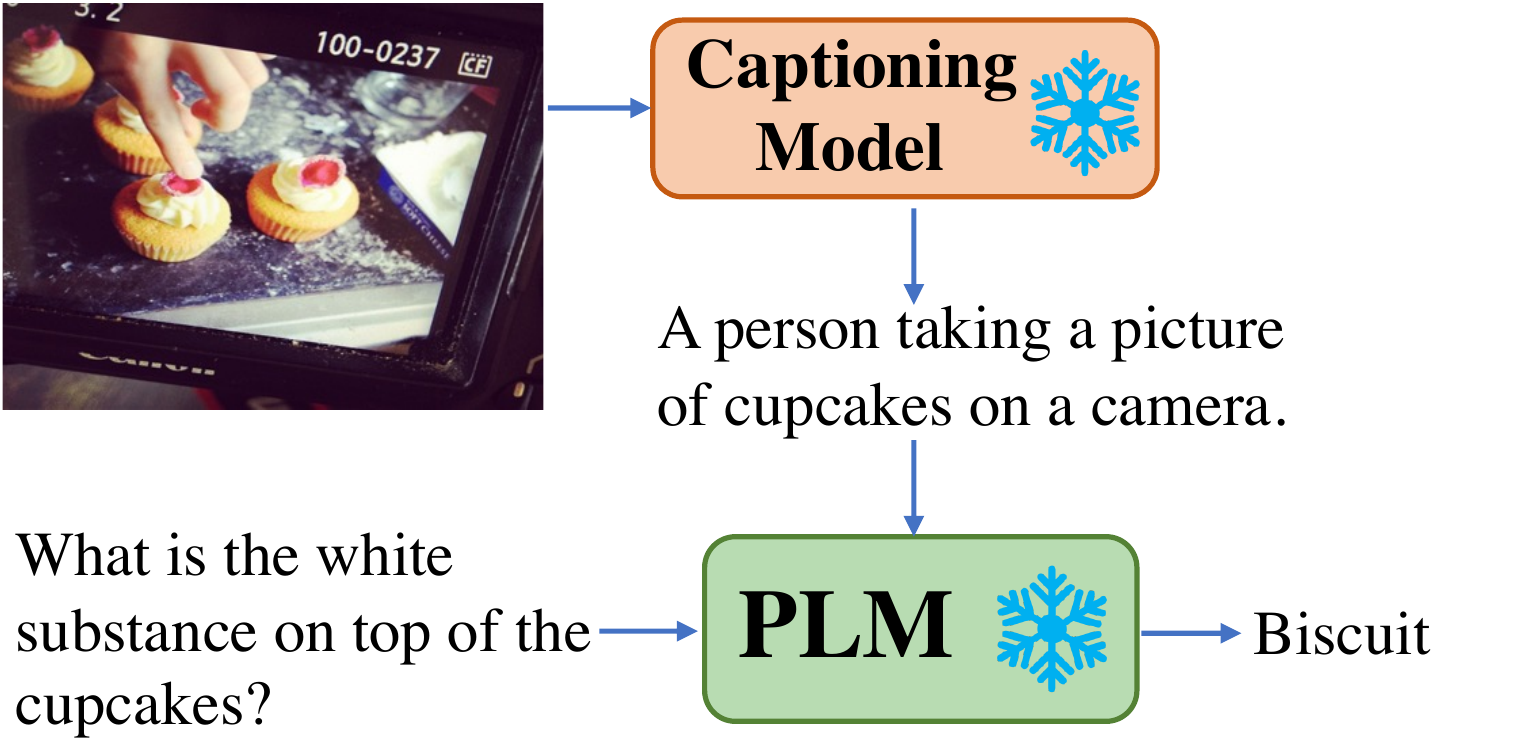}
    \caption{An example that a captioning model (BLIP) fails to provide suitable descriptions for a prediction model (FLAN-T5) of a question in A-OKVQA dataset.}
    \label{fig-intro}
\end{figure}
Inspired by these successes, recent studies have proposed utilizing PLMs\footnote{In this paper, PLMs refer to the models trained on text-only corpus, instead of the text encoder/decoder in vision-language pre-trained~(VLP) models, which typically have a weaker reasoning capacity in linguistic content.}  to solve complex vision-language tasks, exemplified  by the task of knowledge-based \emph{visual question answering}~(VQA) that aims to answer open-ended questions given an image based on outside knowledge~\cite{schwenk2022okvqa}. It has been shown that PLM-enhanced approaches~\citep{gui2021kat, lin2022revive} typically lead to an improved performance on the knowledge-based VQA task than pure vision-language pre-trained (VLP) models~\cite{schwenk2022okvqa}.  

In the literature, existing PLM-enhanced VQA approaches can be roughly categorized into two lines. The first line of research focuses on adapting PLMs to the vision modality by introducing specific modular networks or training objectives~\citep{tsimpoukelli2021multimodal, liang2022modular, alayrac2022flamingo}.
However, they usually incur a high computational cost during pre-training in order to effectively integrate a vision encoder into the PLM. 
As another line of research, several studies aim to reduce the cost of tuning PLMs in vision-language tasks by utilizing PLMs in a zero-shot or few-shot manner. 
They typically generate a caption for an image using  a captioning model~(\eg a fine-tuned VLP model), and employ the generated caption as the context (\eg prompt) to assist PLMs in question answering~\citep{yang2022empirical, tiong2022plug, guo2022images}. 
Such an approach is training-free and can be generally applied with various PLMs.

\ignore{ of modular networks or specialized training objectives to adapt PLMs to the vision modality~\citep{tsimpoukelli2021multimodal, liang2022modular, alayrac2022flamingo}. 
However, this approach typically incurs significant costs in the pre-training stage in order to effectively integrate a vision encoder into the PLM. An alternative approach aims to reduce the cost of deploying PLMs in vision-language tasks by utilizing them in a zero-shot or few-shot manner. 
For instance, \citet{yang2022empirical, tiong2022plug, guo2022images} first convert an image to a caption using a captioning model~(usually a fine-tuned VLP model) and then pass the caption to a PLM as context for question answering. This method is training-free and can be easily generalized to any new PLM.
}

However, in these existing zero-shot or few-shot methods, the captioning model is unaware of both \emph{task goal} and \emph{information need} for the integrated PLM. They directly reuse the captioning model fine-tuned on caption datasets. As a result, the generated captions tend to be less informative for the VQA task, even irrelevant to the question. Figure~\ref{fig-intro} presents an example that an inappropriate caption leads to an incorrect answer generated by the PLM. As we can see, the question is highly related to keywords ``\emph{icing}'' or ``\emph{frosting}'', while the captioning model misses these information and generates a general description. 

\ignore{However, it should be noted that these zero-shot or few-shot methods address the problem in a single-pass paradigm and have a significant drawback. As shown in Figure~\ref{fig-intro}, to answer the question regarding the white substance on top of the cupcakes, the VLP model should generate a caption that includes relevant information such as ``icing'' or ``frosting''. However, it produces a general caption of the image without useful information, which leads to the PLM providing an incorrect answer. The main problem here is the lack of feedback to indicate to the VLP model whether the generated captions are good or not.
}

To address this issue, \ignore{in this paper,}we propose \textbf{\textsc{Lamoc}}: a novel \textbf{\underline{LA}}nguage \textbf{\underline{MO}}del guided \textbf{\underline{C}}aptioning approach for the VQA task. The key idea is to leverage the guidance and feedback of the prediction model (\ie the PLM) to improve the capability of the captioning model, so that it can be aware of the task goal and information need, and assist the prediction model in answer prediction. Our approach is specially designed with two gradual training stages. At the first stage, the captioning model is trained to align to the prediction model, in which the prediction model selects captions that are more pertinent to a given question from multiple propositions generated by the captioning model. These selected captions are informative and can be used to fine-tune the captioning model to generate informative captions. At the second stage, since the generated caption is used by the PLM as direct evidence for VQA, we employ the feedback from the PLM as reward signals to train the captioning model via reinforcement learning. 
During training, only the captioning model is tuned while the PLM is fixed, which significantly reduces the computational costs. Meanwhile, since the feedback is from PLM, both training stages do not require any labeled data.

Our contributions can be summarized as follows: (1) We propose \textbf{\textsc{Lamoc}}, a
novel approach for training captioning models to generate informative captions that can assist PLMs in VQA tasks; 
(2) Using a small number of randomly sampled unlabeled (image, question) pairs, \textbf{\textsc{Lamoc}} consistently outperforms several competitive zero/few-shot baselines without PLM feedback on two knowledge-based VQA datasets: OK-VQA and A-OKVQA; (3) We have demonstrated the effectiveness of our method on PLMs of varying scales, from 223M to 11B. This not only confirms the robustness of our approach but also demonstrates its potential for generalization to Large Language Models~(LLMs).

\section{Related Work}
\paragraph{PLMs for VQA.} After training on large corpora, PLMs exhibit surprising abilities, such as chain-of-thought reasoning~\citep{wei2022chain}, in-context learning~\citep{brown2020language}, and instruction following~\citep{chung2022scaling}, which cannot be obtained by vision-language pre-training. Thus, some works adopt PLM to perform VQA and obtain promising results. One line of research combines a PLM and a vision encoder and trains them end-to-end. Frozen~\citep{tsimpoukelli2021multimodal} and \citet{liang2022modular} train a visual encoder or a modular network and keep the PLM frozen to retain its powerful abilities. 
Flamingo~\citep{alayrac2022flamingo} elaborates the model architecture to combine the vision and language models and scales the model size to 80B. Another line of research tries to deploy PLMs on VQA tasks in a few-shot/zero-shot manner. PICa~\citep{yang2022empirical} and  Img2Prompt~\citep{guo2022images} translate the image to captions or tags and employ GPT-3 to answer a question by in-context learning. PNP-VQA~\citep{tiong2022plug} generates question-related captions and utilizes a QA model~\citep{kha2022unifiedqa} for answer prediction. This type of work does not require extra training and can be adapted to new PLMs. \ignore{Img2Prompt~\citep{guo2022images} translates the image into QA pairs and feeds them to PLM as in-context examples.} Our work follows the second paradigm and is an extension of these works.

\paragraph{Learning from Feedback.} A regular paradigm to train a model is defining a loss function and optimizing it. However, certain objectives, such as coherence, diversity, and toxicity in text generation, may not be easily incorporated into the loss function and learned in an end-to-end manner~\citep{paulus2018rlsum, pang2021text}. Thus, explicit feedback on model output is regarded as a learning signal to assist in training. \citet{campos2022training} utilize a PLM's refinement and human feedback to fine-tune a summary model. \citet{wang2022compilable} leverage compiler feedback to improve the compilability of programs generated by the language model. \citet{ouyang2022training} align a language model with the user’s intention through reinforcement learning from human feedback. We borrow idea from these works, but our feedback comes from a PLM instead of humans, thus saving the annotation cost.
\section{Method}
\begin{figure*}[h]
    \centering
    \includegraphics[width=\textwidth]{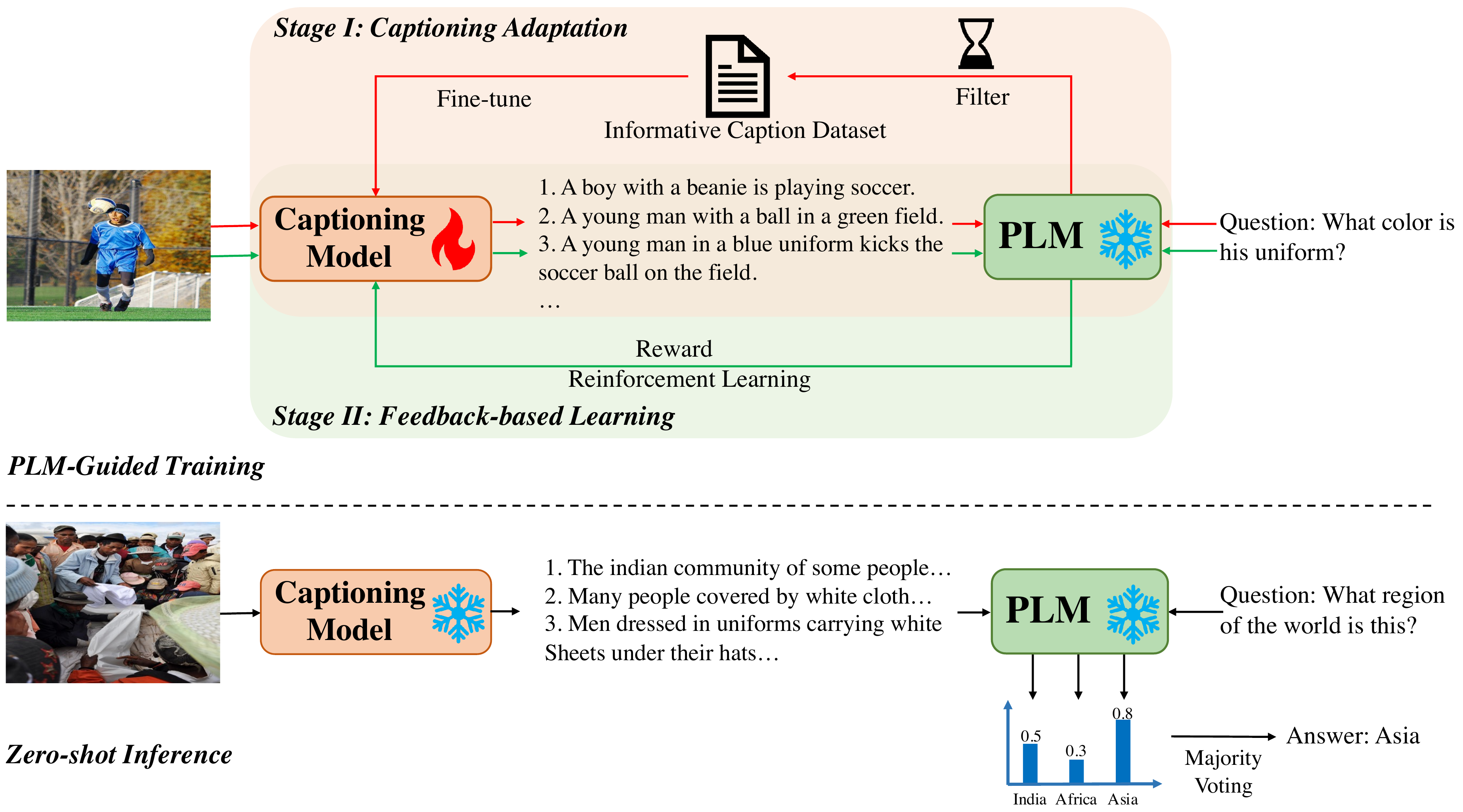}
    \caption{Overview  of our proposed approach \textsc{Lamoc}. In captioning adaption, we utilize a PLM to select informative captions and fine-tune the captioning model on them. When learning from PLM feedback, we regard the feedback from the PLM as reward signals and perform reinforcement learning on the captioning model. \ignore{We utilize BLIP~\citep{li2022blip} as the captioning model and FLAN-T5-XXL~\citep{chung2022scaling} as the PLM.}}
    \label{fig-model}
\end{figure*}

In this section, we present the proposed \textsc{Lamoc}: \textbf{\underline{LA}}nguage \textbf{\underline{MO}}del guided \textbf{\underline{C}}aptioning method for VQA.
The overall architecture of \textsc{Lamoc} is depicted in Figure~\ref{fig-model}.  


\subsection{Overview of Our Approach}

In this work, we study the task of \emph{visual question answering~(VQA)}. Given 
an image-question pair $x: \langle x_i, x_q \rangle$, the task goal is to predict a correct answer $y$ to the question $x_q$ given the image $x_i$. Following prior studies~\cite{yang2022empirical, tiong2022plug}, we adopt a captioning-based approach for VQA, in which a captioning model generates auxiliary captions for helping answer prediction. 
Formally, we represent the above idea in a probabilistic way: 
\begin{eqnarray}\label{eq-piq}
&&p(y|x_{i}, x_{q}) \\ 
&= & \sum_{z \in \mathcal{Z}} \underbrace{p(z|x_{i}, x_q; \Theta_{C})}_{\text{caption generation}} \cdot \underbrace{p(y|x_{q},z; \Theta_{P})}_{\text{answer prediction}}, \nonumber
\end{eqnarray}
where a captioning model $\Theta_{C}$ firstly generates an auxiliary captions $z$, and then 
a prediction model $\Theta_{P}$ predicts an answer candidate $y$ based on  the caption $z$ and the question $x_q$. We evaluate this probability by iterating over a set of generated captions. 
Here, we consider an unsupervised setting: no labeled answer data is available. Although there is no labeled answers, we assume that a small number of 
image-question pairs can be obtained for training (no overlapping with the task dataset).

To instantiate this probabilistic approach, we adopt a  vision-language pre-trained~(VLP) model, \ie BLIP~\cite{li2022blip}, as the captioning model, and a pre-trained language model~(PLM), \ie FLAN-T5-XXL~\cite{chung2022scaling}, as the prediction model. 
The prediction model $\Theta_{P}$ is expected to fulfill the task by accurately predicting the answer, while the captioning model $\Theta_{C}$ plays an assisted role by providing  informative evidence for $\Theta_{P}$. 
In our approach, the captioning model $\Theta_{C}$ can be tuned while the  prediction model $\Theta_{P}$ is fixed during optimization. By leveraging the unlabeled image-question pairs (without the labeled answers), we let the two models  cooperate with each other: the captioning model generates  informative evidence  for helping answer prediction, and the prediction model provides task-specific guidance and feedback to improve the captioning model. 

To optimize our approach, we design a gradual training process including two stages: 
(1) \emph{captioning adaptation}  aims to adjust $\Theta_{C}$ to produce informative captions that are suitable for $\Theta_{P}$~(\S\ref{cap_gen}), and (2)  \emph{feedback-based learning} aims to optimize  $\Theta_{C}$ according to task-specific feedback from $\Theta_{P}$~(\S\ref{rl}). 
Once the captioning model is well trained, we employ the prediction model for predicting the final answer as in Eq.~\eqref{eq-piq}, based on the captions provided by the captioning model~(\S\ref{inference}). Next,  we introduce these parts in details.

\subsection{Language Model Guided Captioning}

The key of our approach (Eq.~\eqref{eq-piq}) is to train an effective captioning model $\Theta_C$ for improving the capability of the prediction model $\Theta_P$ on VQA. Considering that  there are no labeled answers, we employ  the prediction model to provide guidance and feedback to optimize  the captioning model. 

\subsubsection{Captioning Adaptation}\label{cap_gen}

Since the captioning model is originally intended to describe the given image, it may not be in suited form to assist the prediction model. Thus, we propose a captioning adaptation strategy that tunes the captioning model to fit the prediction model.  


\paratitle{Caption Propositions.} We first sample $n$ image-question pairs from VQAv2~\citep{goyal2017making}, which is a large VQA dataset containing more than $1$M questions and does not overlap with our task dataset. Then we employ the captioning model to propose $k$ captions for each image by nucleus sampling~\cite{holtzman2019curious}. Among these captions, some may be better suited for the prediction model than the rest. We would like to identify such captions and use them to refine the captioning model.  

\paratitle{Instruction-based Captions Selection.} 
Since the prediction model is developed based on the FLAN-T5-XXL, it has encoded a large amount of knowledge in a massive number of parameters. 
We design the following instruction to prompt FLAN-T5-XXL to identify more informative captions: 

\emph{``Question: \texttt{[QUESTION]} Caption: \texttt{[CAPTION]}\textbackslash n To what degree does the caption relate to the question:\textbackslash n A: 0\%\textbackslash n B: 25\%\textbackslash n C: 50\%\textbackslash n D:75\%''.} 

\noindent Given the above prompt, FLAN-T5-XXL will generate a corresponding option among the set $\{A, B, C, D\}$. Such an option reflects the correlation between the caption and question, and the captions with the predicted  option 
``\emph{D:75\%}'' are more relevant to the question. 
Since the options are made by the prediction model itself, they tend to be more useful for answer prediction. Thus, we keep the captions with the  predicted option ``\emph{D:75\%}'' and discard the rest.

\ignore{If this score could truthfully reflect the correlation between the caption and question, we could select the captions assigned with option ``D:75\%'' as the adaptation-training examples. 
For an image $x_q$, we select from the group with  ``D:75\%''
}



\paratitle{Captioning Model Fine-tuning.} Via the above caption selection, we can obtain a set of more informative captions, which are judged by the prediction model. Further, we use them to fine-tune the captioning model by optimizing the following cross-entropy loss:

\ignore{where $T$ is the length of caption, $z_{t,j}^*$ denotes the label of the $t^{\text{th}}$ token of the informative caption selected by FLAN-T5-XXL in class $j$, $\hat{z}_{t,j}$ is the $t^{\text{th}}$ token predicted by the captioning model in class $j$.}
\begin{equation}
    \mathcal{L}_{FT} = -\frac{1}{T}\sum\limits_{t=1}^T \log p(z_t|x_i, z_{<t}),
\end{equation}
where $T$ is the length of caption, $z_t$ denotes the $t$-th token of the informative caption selected by FLAN-T5-XXL, $z_{<t}$ represents the generated token up to the $(t-1)$-th step. After fine-tuning, the captioning model can be better suited for  the prediction model. 

\subsubsection{Feedback-based Learning}\label{rl}
Though adapting to the prediction model, the captioning model is still unaware of the answer prediction task for VQA.  Thus, we further propose construct pseudo supervision signals based on the PLM feedback from the prediction model. Since the captioning model is only involved as an intermediate component for answer prediction, we design a reinforcement learning method for optimizing it.


\paratitle{Reward From PLM Feedback.} A key design consideration of reinforcement learning is the definition of the reward function. In our approach, instead of only generating relevant captions for the images, the effectiveness of the captioning model should be measured by how well it helps find the correct answer. To achieve this goal, we design the following two kinds of reward signals.

\textbullet~\emph{Prompt-based Reward:} A heuristic method is utilizing the prompt in \S\ref{cap_gen} to instruct FLAN-T5-XXL to obtain a relevance score, and regard this relevance score as the reward signal:
\begin{equation}
r(x_q, z) = \mathop{\arg\max}\limits_{s\in\{0,0.25,0.5,0.75\}} p(s|x_{q}, z; \Theta_{P}), 
\end{equation}
\ignore{where $f$ is a function maps $\{A, B, C, D\}$ to $\{0, 0.25, 0.50, 0.75\}$.} 
\ignore{Then $\{A,B,C,D\}$ is converted to $\{0,0.25,0.5,0.75\}$.} A higher score indicates a more informative caption, which is encouraged.

\textbullet~\emph{Confidence-based Reward:} Since there is no ground-truth answer during training, following Eq.\eqref{eq-piq}, we employ the probability score of the predicted answer (the most confident candidate) given by the prediction model as the reward:
\begin{equation}
r(x_q, z) = p(\hat{y} | x_{q}, z; \Theta_{P}), 
\end{equation}
where $z$ is the generated caption by the captioning model and $\hat{y}$ is the predicted answer from the prediction model. 
In this way, the PLM (\ie the prediction model) can inform the captioning model about the informativeness of the generated caption: the larger probability score,  the more informative a caption is, and vice versa. We will verify the reliability of these reward designs in \S\ref{reliable}. 

\paratitle{Policy Gradient.} In the framework of reinforcement learning, caption generation can be viewed as a sequential decision-making process over the whole vocabulary space. Each generated caption with $T$ tokens is treated as an individual episode of length $T$ in this process. At the $t$-th time step, the state $(x_i, z_{<t})$ is the combination of the image and caption generated up to the $(t-1)$-th token, and the action $z_t$ is the $t$-th token to be generated. We employ the policy gradient algorithm~\citep{sutton2018reinforcement} and perform gradient descent to optimize the following objective function:
\begin{equation}
    \mathcal{L}_{RL} = -\sum\limits_{t=1}^{T} r(x_q, z)\log p(z_t|x_i, z_{<t}; \Theta_{cap}), 
\end{equation}
where $z= \langle z_1, ..., z_t, ..., z_T \rangle$ is the caption, and $r(x_q, z)$ is the reward given by the PLM. 
\ignore{In the implementation stage, we minus $r(x_q, z)$ by the mean of reward in a mini-batch and take the negative of the objective function $R$ as the final loss function:
\begin{equation}
    \mathcal{L}_{RL}=-R.
\end{equation}
}
Finally, we jointly optimize the two loss functions:
\begin{equation}
    \mathcal{L}=(1-\alpha)\cdot \mathcal{L}_{FT}+\alpha\cdot\mathcal{L}_{RL},
\end{equation}
where $\alpha$ is a weight factor to balance the two parts. 

To fully exploit the online feedback provided by FLAN-T5-XXL, we only optimize the captioning adaptation loss function $\mathcal{L}_{FT}$ in the initial epoch, while the reinforcement learning loss function $\mathcal{L}_{RL}$ is optimized throughout the training process.

\subsection{Answer Prediction}\label{inference}
At inference time, we utilize the updated captioning model to assist the prediction model in answering questions, by calculating the probability $p(y|x_{q},z; \Theta_{P})$. 
To increase the diversity of captions and the coverage of answers, we first randomly sample $20\%$ patches from the whole image at each time and apply top-$k$ sampling~\citep{fan2018hierarchical} to generate a caption for these patches with the updated captioning model. We repeat this process $m$ times to generate $m$ diverse captions. Then we concatenate each of them with the corresponding question to construct the following prompt: 

\emph{``Please answer the following question.\textbackslash n\texttt{[CAPTION]}. \texttt{[QUESTION]}''.} 

\noindent Based on this prompt,  the FLAN-T5-XXL is instructed to propose an answer with greedy decoding. We can  take the max-voting strategy over all the generated answers. \ignore{or top three answers with the highest probability.} 

Different from previous work on learning from feedback~\cite{campos2022training, wang2022compilable, ouyang2022training}, our proposed approach explores the guidance and feedback from the prediction model instead of human annotations.  As we will see in \S\ref{reliable}, our empirical study shows that there exists a negative correlation between the negative log likelihood assigned by a PLM and the VQA score of a generated answer. This finding suggests that the reward $r(x_q, z)$ given by PLM can potentially serve as a substitute for labeled data to improve the captioning model for the VQA task. \ignore{Another important design is that the captioning adaptation strategy that adapts the captioning model to the prediction model  (\textcolor{blue}{Eq. or Sec???}), in order to cooperate in a more effective way, which is verified in \S\ref{two-stage-ablation}.}

\section{Experiment}
\begin{table*}[ht]
\centering
\small
\begin{tabular}{ccccccc}
\toprule

\makecell[c]{Evaluation\\ Setting} & Method & Parameters & \makecell[c]{Use\\ Extra PLM?} & \makecell[c]{With extra V-L\\Pre-training?} & \makecell[c]{OK-VQA\\test} & \makecell[c]{A-OKVQA\\val}\\
\midrule
\multicolumn{7}{l} {\textbf{Models fine-tuned on training set.}} \\[1ex]
\multirow{2}{*}{\makecell[c]{Supervised \\ learning}} & BLIP$^\dag$ & 226M & \ding{55} & \ding{55} & 37.6 & 38.5 \\
 & \makecell[c]{PromptCap} & 175B & \ding{52} & \ding{55} & 58.8 & 58.0 \\
\midrule
\multicolumn{7}{l} {\textbf{Models without fine-tuning.}} \\[1ex]
\multirow{3}{*}{Few-shot} & FewVLM\textsubscript{base} & 288M & \ding{55} & \ding{52} & 15.0 & - \\
 & FewVLM\textsubscript{large} & 804M & \ding{55} & \ding{52} & 23.1 & - \\
 & PICa  & 175B & \ding{52} & \ding{55} & 48.0 &  - \\
\cmidrule{2-7}
\multirow{12}{*}{Zero-shot} & VL-T5\textsubscript{no-VQA} & 288M & \ding{55} & \ding{52} & 5.8 & - \\
 & VLKD\textsubscript{ViT-B/16} & 494M & \ding{52} & \ding{52} & 10.5 & - \\
 & VLKD\textsubscript{ViT-B-L/14} & 713M & \ding{52} & \ding{52} & 13.3 & - \\
 & Flamingo\textsubscript{3B} & 3B & \ding{52} & \ding{52} & 41.2 & -\\
 & Flamingo\textsubscript{9B} & 9B & \ding{52} & \ding{52} & 44.7 & -\\
 & Flamingo\textsubscript{80B} & 80B & \ding{52} & \ding{52} & 50.6 & -\\
 & Frozen & 7B & \ding{52} & \ding{52} & 5.9 & - \\
 & PNP-VQA\textsubscript{3B}  & 3.9B & \ding{52} & \ding{55} & 34.1 & 33.4 \\
 & PNP-VQA\textsubscript{11B}  & 11.9B & \ding{52} & \ding{55} & 35.9 & 36.0 \\
 & Img2Prompt\textsubscript{6.7B} & 8.3B & \ding{52} & \ding{55} & 38.2 & 33.3 \\
 & Img2Prompt\textsubscript{13B} & 14.6B & \ding{52} & \ding{55} & 39.9 & 33.3 \\
 & \textbf{\textsc{Lamoc}\textsubscript{11B}~(Ours)}  & 11.4B & \ding{52} & \ding{55} & 40.3 & 37.9 \\
\bottomrule
\end{tabular}
\caption{\label{main_result}
Results on OK-VQA and A-OKVQA. The methods are categorized by whether they use extra PLM and whether carry out V-L pre-training. The methods in the upper part have been fine-tuned on the training set, while those in the middle and bottom parts have not. All methods using extra PLM keep it frozen. $^\dag$ Instead of first fine-tuning BLIP on VQAv2 and then performing task-specific fine-tuning, we directly fine-tune BLIP on two target datasets for a fair comparison.
}
\end{table*}
This section shows the experimental setup and then highlights the main conclusions of our results.

\subsection{Experimental Setup}
\paratitle{Task Datasets.}
Since our goal is to improve the performance of PLMs on visual commonsense tasks, we choose two knowledge-based VQA datasets to evaluate our method: (1) \textbf{OK-VQA}~\citep{marino2019ok} contains 5,046 questions in the test set that require external knowledge resources to answer. (2) \textbf{A-OKVQA}~\citep{schwenk2022okvqa} is an augmented dataset based on OK-VQA, which requires additional types of world knowledge compared to OK-VQA. Since the test set of A-OKVQA is not public, we evaluate our method on the validation set. We do not test on VQAv2~\citep{goyal2017making} because the majority of questions in this dataset are largely focused on recognition and simple visual detection tasks, which can be done without much logical reasoning or external knowledge, and a fine-tuned VLP model could obtain surprising results~\cite{wang2022image, wang2022ofa}. 
We do not use training data to make a fair comparison with other methods. 

\paratitle{Baselines.}
We divide previous methods into two categories: (1) \textbf{Methods without extra large-scale Vision-Language~(V-L) pre-training}, which means the models have not been pre-trained on large-scale V-L datasets, including PICa~\citep{yang2022empirical}, PNP-VQA~\citep{tiong2022plug}, Img2Prompt~\citep{guo2022images}. \textsc{Lamoc} also belongs to this category. (2) \textbf{Methods with extra large-scale V-L pre-training}, which means that the PLM and the vision encoder are jointly trained on V-L datasets~(although the PLM may be fixed, it obtains the ability to understand images), including VL-T5~\citep{cho2021unifying}, FewVLM~\citep{jin2021good}, VLKD~\citep{dai2022enabling}, Frozen~\citep{tsimpoukelli2021multimodal}, and Flamingo~\citep{alayrac2022flamingo}. 
The above methods do not use or use few labeled data (zero-shot/few-shot).
Besides, we include two methods, \ie BLIP~\cite{li2022blip} and PromptCap~\cite{hu2022promptcap}, which are fine-tuned on large amounts of labeled data.

\paratitle{Implementation details.} 
For image captioning, we adopt BLIP~\citep{li2022blip} with 446M parameters and load the released checkpoint that has been fine-tuned on the COCO 2014 training set~\citep{lin2014microsoft}, which has no overlap with both the OK-VQA and A-OKVQA evaluation datasets. For the PLM, we utilize FLAN-T5-XXL~\citep{wei2022chain}, which has been fine-tuned on more than 1,800 tasks through instructions and stores considerable world knowledge. We also carry out experiments on PLMs with other sizes, from 223M to 11B parameters, to demonstrate the robustness and generalizability of our approach across PLMs with different sizes. It is noteworthy that the informative caption dataset used in the captioning adaptation stage is selected by FLAN-T5-XXL, because the relevance score given by smaller models is not reliable, as will be illustrated in \S\ref{reliable}. When training the captioning model, we select 1,000 (image, question) pairs without labels from VQAv2 (about $10\%$ of the amount of training data for our target datasets), which has no overlap with the OK-VQA and A-OKVQA. It is worth noting that these 1,000 image-question pairs can be sampled from any datasets or even be generated, we sample from VQAv2 for the sake of reproducibility. The answers are generated by the PLM auto-regressively, without access to the pre-defined answer list. We conduct experiments with 5 random seeds and report the average VQA score according to official evaluation protocols.

\subsection{Main Results}
\ignore{\begin{table*}
\centering
\begin{tabular}{lcccc}
\toprule
Method & FLAN-T5-base & FLAN-T5-large & FLAN-T5-XL & FLAN-T5-XXL \\
\midrule
BLIP caption & 19.46 & 28.86 & 31.21 & 33.9 \\
+ adapt & 18.71 & 29.19 & 31.07 & 35.20  \\
+ rl (prompt) & 19.25 & 29.73 & 30.63 & \textbf{38.17} \\
+ rl (confidence) & 19.74 & 28.98 & \textbf{32.10} & -\\
+ adapt + rl & \textbf{20.63} & \textbf{29.84} & 32.00 &  37.63 \\
\bottomrule
\end{tabular}
\caption{\label{stage_ablation}
Results of different model scale and different training objectives. ``BLIP caption'' means there is no language model guided training involved. ``adapt'' means captioning adaptation while ``rl'' means reinforcement learning from PLM feedback.
}
\end{table*}}

Table~\ref{main_result} displays the results of our methods and baselines on OK-VQA and A-OKVQA. 

First, \textsc{Lamoc} outperforms PNP-VQA on both datasets, a strong zero-shot baseline without large-scale V-L pre-training. Compared to PNP-VQA and Img2Prompt with similar model sizes, \textsc{Lamoc} achieves prominent gains on the challenging A-OKVQA dataset (38.2 vs 36.0 and 38.2 vs 33.3). These significant achievements clearly show the effectiveness of our method. However, the improvements on OK-VQA are marginal. This is because questions in OK-VQA do not require too much reasoning~\cite{schwenk2022okvqa}, and the answers are likely to be contained in general captions. Therefore, some baselines without training from feedback also achieve comparable or even better results.
\ignore{\textcolor{blue}{The performance of \textsc{Lamoc} on OK-VQA does not surpass that of Img2Prompt. This can be attributed to the utilization of a different PLM for answer prediction in our method, and the absence of an image-question matching model and a question generation model which are critical components of Img2Prompt. We will show in Table~\ref{stage_ablation} that our method improves the performance of FLAN-T5 on OK-VQA}.}
While, since Flamingo has been trained on a massive V-L dataset, it achieves the best performance among zero-shot methods. It has been reported that large-scale V-L pretraining can develop a mapping between images and knowledge concepts that can aid in knowledge-based VQA~\citep{tiong2022plug}. Compared to these baselines, our approach does not require additional image-question matching or question generation modules, thus speeding up the inference speed. 


Second, \textsc{Lamoc} narrows the gap between methods with and without fine-tuning, and even achieves comparable results with the fine-tuned VLP model, \ie BLIP. For example, the performance gap between PNP-VQA\textsubscript{11B} and BLIP is 2.5, and has been decreased to 0.3 by \textsc{Lamoc}, which implies the importance of language model feedback.

Finally, we report the results of our methods with different model sizes in Table~\ref{stage_ablation}. When increasing the model scale from 223M to 11B, we observe a 1-4 point improvement in VQA scores on the challenging A-OKVQA dataset. This indicates that a larger PLM can not only store more world knowledge to assist with question answering, but also provide more accurate feedback to refine the captioning model. This is further supported by the ablation study in \S\ref{reliable}.


\section{Analysis}
\subsection{The Reliability of Feedback From PLM}\label{reliable}
\begin{figure}[ht]
    \centering
    \setlength{\abovecaptionskip}{0.cm}
    \includegraphics[width=0.5\textwidth]{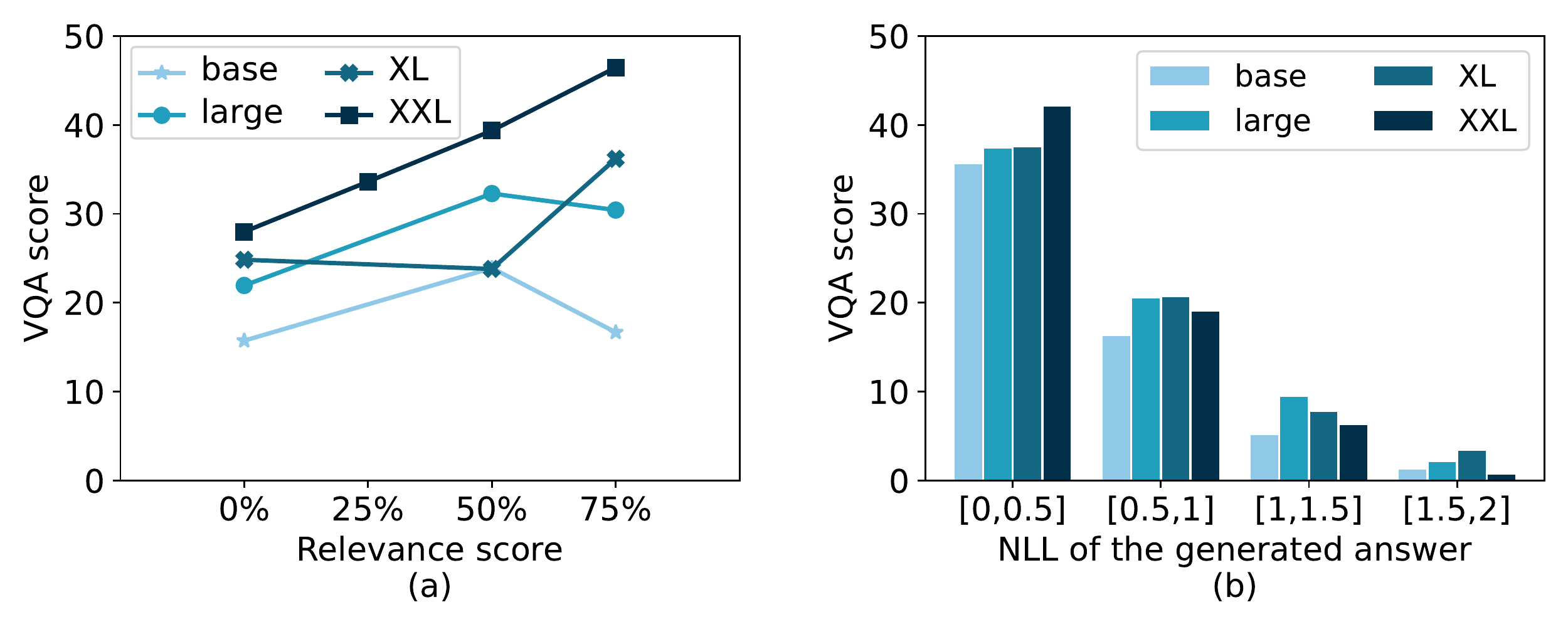}
    \caption{The relationship between the caption's reward and the corresponding answer's VQA score on A-OKVQA validation set. Figure (a) reflects the reliability of prompt-based reward while figure (b) reflects the reliability of confidence-based reward.}
    \label{fig-reliable}
\end{figure}
The main idea of our work is leveraging the feedback of a PLM to guide caption generation, so a critical aspect is the reliability of the feedback. \textsc{Lamoc} involves two types of feedback: (1) prompt-based reward and (2) confidence-based reward, which will be evaluated independently. 

To evaluate the reliability of the first type of feedback, we analyze the relation between the VQA score and the relevance score provided by the PLM on A-OKVQA validation set~(Figure~\ref{fig-reliable}(a)). We can observe that as the relevance score provided by FLAN-T5-XXL increases, the VQA score also increases, indicating that FLAN-T5-XXL is a suitable prediction model for providing accurate feedback and the relevance scores can be regarded as reward signals. However, this trend is not observed for the other three models, implying that their feedback is unreliable. As a result, we only use FLAN-T5-XXL to select informative captions during captioning adaptation. 

To evaluate the reliability of the second type of feedback, we prompt FLAN-T5 to answer the question conditioned on the captions and plot the relationship between the negative log-likelihood (NLL) of the generated answer and its corresponding VQA score. As Figure~\ref{fig-reliable}(b) shows, there is a negative correlation between the NLL of the generated answers and their VQA scores, suggesting that captions with lower NLL are more informative and relevant to the questions. Therefore, the probability of the generated answer is a reliable feedback and can be used as the reward signal during reinforcement learning.

\subsection{The Effectiveness of Two-stage Training}\label{two-stage-ablation}

\begin{table*}
\small
\centering
\renewcommand\arraystretch{1.1}
\setlength\tabcolsep{4pt}
\begin{tabular}{lcccccccc}
\toprule
\multirow{2}{*}{Method} & \multicolumn{2}{c}{FLAN-T5-base (223M)} & \multicolumn{2}{c}{FLAN-T5-large (738M)} & \multicolumn{2}{c}{FLAN-T5-XL (3B)} & \multicolumn{2}{c}{FLAN-T5-XXL (11B)} \\
 & OK-VQA & A-OKVQA & OK-VQA & A-OKVQA & OK-VQA & A-OKVQA & OK-VQA & A-OKVQA \\
\midrule
BLIP caption & 20.42 & 19.46 & 23.86 & 28.86 & 32.36 & 31.21 & 38.48 & 35.06 \\
+ adaptation & 19.72 & 18.71 & \textbf{27.43} & 29.19 & 32.22 & 31.07 & 38.35 & 35.30 \\
+ RL (prompt) & \textbf{21.24} & 19.25 & 27.29 & 29.73  & 32.28 & 30.63 & 38.74 & 37.62 \\
+ RL (confidence) & 21.14 & 19.74 & 25.09 & 28.98 & 32.02 & \textbf{32.10} & \textbf{40.31} & \textbf{37.85} \\
+ adaptation + RL & 19.72 & \textbf{20.63} & 24.82 & \textbf{29.84} & \textbf{32.77} & 32.00 & 39.72 & 37.09 \\
\bottomrule
\end{tabular}
\caption{\label{stage_ablation}
Results of different model sizes and different training objectives. ``BLIP caption'' means feeding captions generated by BLIP to PLM without captioning adaptation and feedback-based learning. ``adaptation'' means captioning adaptation, while ``RL~(prompt)'' means RL with prompt-based reward, and ``RL~(confidence)'' means RL with confidence-based reward.
}
\end{table*}

When training the captioning model, we adopt two gradual training stages: captioning adaptation and feedback-based learning. In this part, we study the effectiveness of this training strategy and explore whether one training stage is more effective than the other. As illustrated in Table~\ref{stage_ablation}, different models benefit from different training objectives. For example, the captioning adaptation stage is more beneficial for FLAN-T5-large, leading to an improvement of about 4 points on OK-VQA. On the other hand, FLAN-T5-XXL benefits the most from reinforcement learning with prompt-based rewards and obtains more than 4 points improvement on A-OKVQA. Moreover, the results show that jointly training the two objectives further boosts performance, highlighting the effectiveness of the proposed two-stage training approach.

\subsection{Case Study}\label{case_study}
\begin{figure*}[ht]
    \centering
    \includegraphics[width=0.9\textwidth]{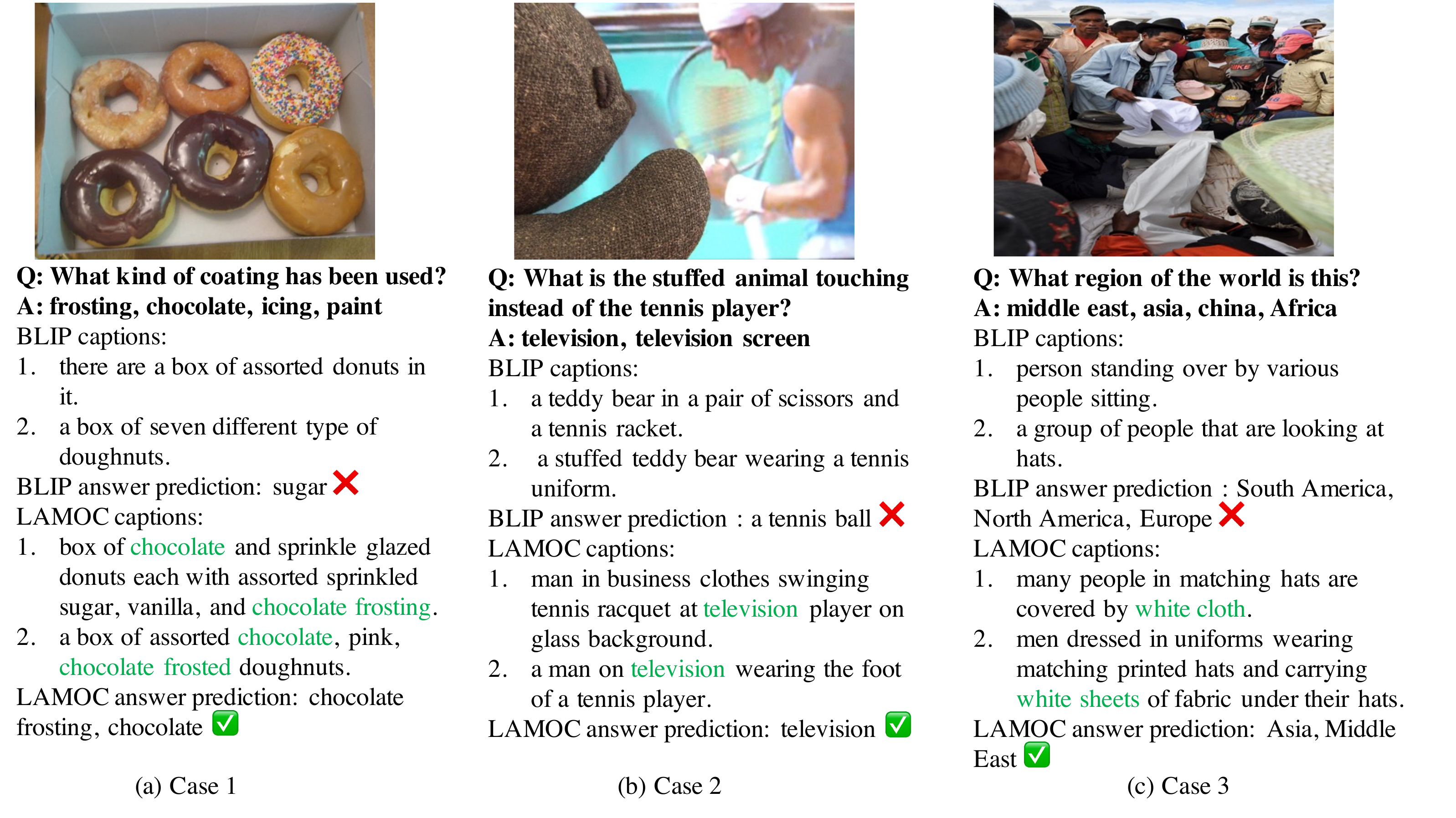}
    \caption{Example captions and predictions generated by BLIP and \textsc{Lamoc}.}
    \label{fig-case}
\end{figure*}
Figure~\ref{fig-case} displays three instances of the captions generated by BLIP and \textsc{Lamoc}, along with the corresponding answers generated by FLAN-T5-XXL. Since \textsc{Lamoc} is trained on the basis of BLIP, the difference can reflect the effect of our method. As can be observed, the captions generated by \textsc{Lamoc} are longer and more comprehensive, containing key information relevant to the question. For example, in Figure~\ref{fig-case}(a), \textsc{Lamoc} generates captions that include specific details such as ``\textit{frosting}'' and ``\textit{chocolate}'', while BLIP only generates general captions about ``\textit{donuts}'' and ``\textit{box}'', without sufficient information to help answer the question. These results highlight the importance of training the captioning model under the guidance of PLMs.

One concern is that the PLM may generate correct answers due to the language bias, not attributing to the relevant information contained in the captions. For example, in Figure~\ref{fig-case}(a), the PLM may generate the answer ``\textit{chocolate}'', even if the captions do not mention chocolate~\cite{li2023evaluating}. However, since chocolate often co-occurs with donuts in the training corpora, the PLM may associate chocolate with donuts and generate it as the answer. In order to check how often such a situation happens, we randomly sample 100 questions where the prediction model gives correct answers. For each question, we manually assess whether their answer is derived from the caption. Our analysis reveals that only 6 out of 100 captions are irrelevant to the questions, indicating the reliability of the captions.

Another interesting phenomenon is that the sentences generated by \textsc{Lamoc} can be grammatically incoherent and sometimes incomplete. This indicates that PLM prompting may not always conform to human language patterns, which is consistent with previous studies~\citep{webson2021prompt, deng2022rlprompt}.

The ablation study of the level of relevance, the number of captions, and the influence of different prompt designs can be found in appendix~\ref{app-ablation}.

\section{Conclusion}
In this paper, we propose \textsc{Lamoc}, a language model guided captioning method that improves a captioning model to generate comprehensive captions for an image to help answer the question. In order to train such a model, we first perform captioning adaptation on a self-generated dataset filtered by FLAN-T5-XXL, and then fine-tune the updated captioning model through reinforcement learning from PLM feedback. Our method, \textsc{Lamoc}, generates captions that are both informative and able to assist PLMs in VQA tasks, as demonstrated through experiments on two knowledge-based VQA datasets. On the challenging A-OKVQA dataset, \textsc{Lamoc} substantially outperforms previous zero-shot methods and even achieves comparable results to a fine-tuned VLP model. Additionally, we show that \textsc{Lamoc} is generalizable to PLMs of varying sizes, from 223M to 11B parameters, demonstrating its potential to be applied to LLMs, which we leave as future work.
\section{Limitations}
In our study, we have demonstrated the effectiveness of our proposed method on FLAN-T5 with different sizes. However, we have not yet evaluated its performance on LLMs, which possess an even greater number of parameters and have been pre-trained on larger corpora, thus potentially providing more accurate feedback for both caption adaptation and reinforcement learning. Meanwhile, it is worth noting that PLMs may contain certain biases, and training based on their feedback may amplify these biases. As future work, we aim to investigate the scalability of our method to LLMs, as well as strategies to mitigate the potential negative effects of biases present in PLMs.
\section*{Ackownledgement}
This work was partially supported by National Natural Science Foundation of China under Grant No. 62222215, Beijing Natural Science Foundation under Grant No. 4222027, and Beijing Outstanding Young Scientist Program under Grant No. BJJWZYJH012019100020098.  Xin Zhao is the corresponding author.
\bibliography{anthology,custom}
\bibliographystyle{acl_natbib}

\clearpage

\appendix

\section*{Appendix}
\section{Training Details and Artifacts}\label{app-exp}
For \textsc{LAMOC} training, we adopt the officially released BLIP captioning checkpoint\footnote{\url{https://storage.googleapis.com/sfr-vision-language-research/BLIP/models/model_large_caption.pth}} for model initialization. For both captioning adaptation and reinforcement learning, we adopt the following hyper-parameters: learning rate $2e-6$, warmup 600 steps, weight decay $0.05$, batch size 8. The balance factor $\alpha$ is set to $0.9$. We train the model for 10 epochs and choose the one with the highest reward~(without labels from the validation set). All the experiments are conducted based on LAVIS~\citep{li2022lavis} under BSD 3-Clause License. The A-OKVQA is under the Apache License 2.0.

\section{Additional Ablation Study}\label{app-ablation}
\subsection{Level of Relevance}
When prompting the PLM to give a correlation score for the caption, the level of relevance is part of the prompt, thus can influence the result. We try different levels for the prompt-based reward and the results are in Table~\ref{tab-level_ablation}. Since four levels gives the highest vqa score, we use four levels in our prompt-based reinforcement learning.

\begin{table}[ht]
\small
\centering
\setlength{\tabcolsep}{1mm}
\begin{tabular}{lc}
\toprule
Level & A-OKVQA\\
\midrule
\makecell[l]{A: 0\%; B: 100\%} & 27.25 \\
\makecell[l]{A: 0\%; B: 50\%; C: 100\%} & 28.29 \\
\makecell[l]{A: 0\%; B: 25\%; C: 50\%; D: 75\%} & 28.98 \\
\makecell[l]{A: 0\%; B: 25\%; C: 50\%; D: 75\%; E: 100\%} & 27.96 \\
\bottomrule
\end{tabular}
\caption{\label{tab-level_ablation}
VQA score of models trained with different levels of prompt-based rewards.
}
\end{table}

\subsection{Number of Captions}
\begin{figure}[ht]
    \centering
    \includegraphics[width=0.5\textwidth]{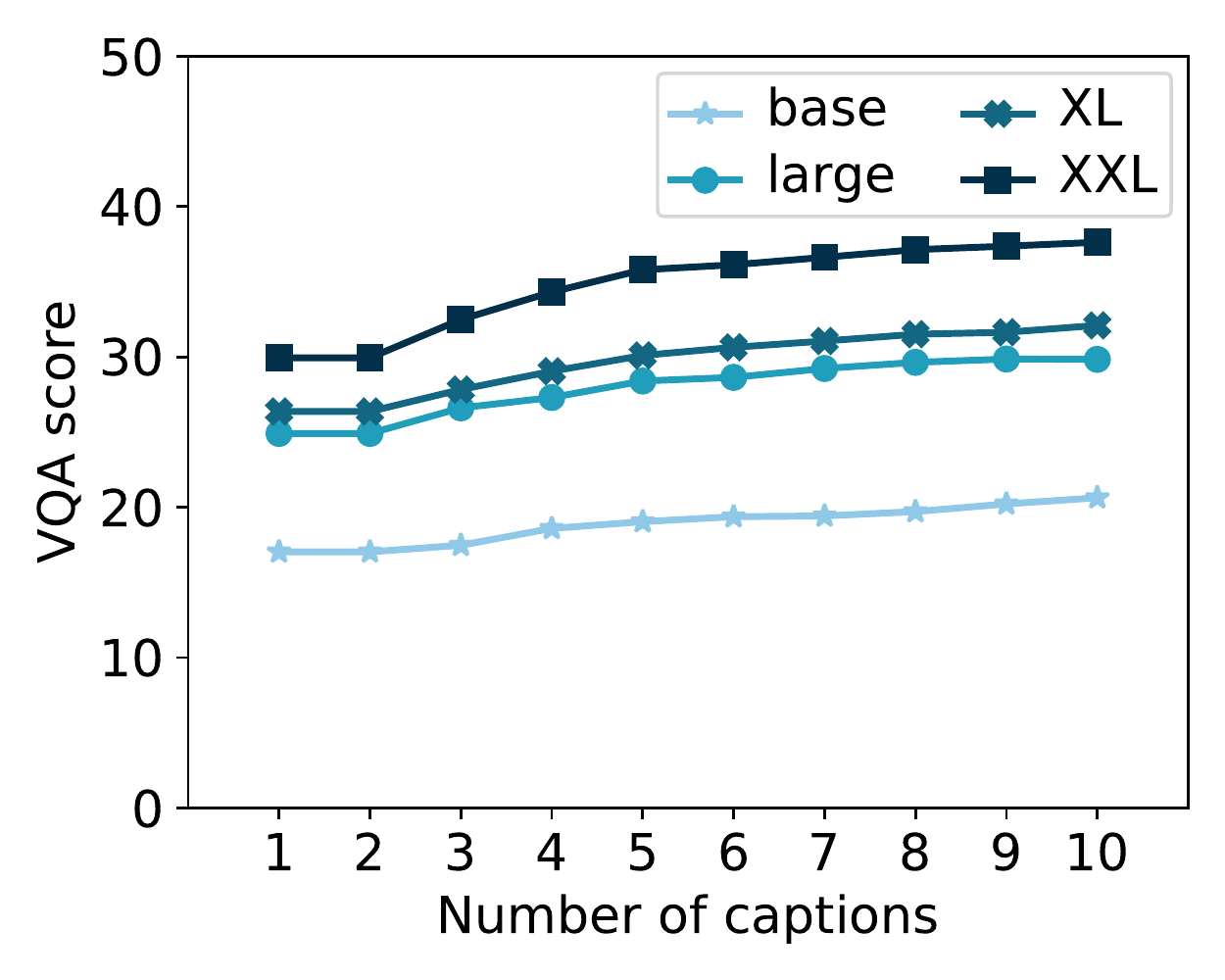}
    \caption{VQA score with different number of captions in the A-OKVQA validation set.}
    \label{fig-caption}
\end{figure}

Since the PLM is "blind," all visual information is carried by the captions. Thus, the number of captions is critical for the PLM to answer the question. In Figure \ref{fig-caption}, we explore the influence of the number of captions. Our results indicate that utilizing a larger number of captions leads to improved performance across various model sizes. Performance gains continue to accumulate even when utilizing 10 captions, leading us to posit that incorporating an even greater number of captions would result in further improvements.

\subsection{Prompt Design}

Another critical design of our method is instructing the FLAN-T5 to provide feedback and answer questions, so we explore the effects of different formats of instruction in Table \ref{tab-prompt_ablation}. We can observe that prompt design has a great impact on the results
Table~\ref{tab-prompt_ablation}, which is in line with the conclusion of previous works~\citep{wei2022chain}. 
\begin{table}[ht]
\small
\centering
\setlength{\tabcolsep}{1mm}
\begin{tabular}{lcc}
\toprule
Prompt & OK-VQA & A-OKVQA\\
\midrule
\makecell[l]{Answer the following question in\\ one word. Q: {[caption]}. {[question]}} & 29.53 & 29.84 \\
\makecell[l]{Please answer the following\\question. {[caption]}. {[question]}} & 28.22 & 29.73 \\
{[caption]}. {[question]} & 27.59 & 27.99 \\
\makecell[l]{{[caption]}. {[question]} Let's think\\ step by step.} & 18.08 & 28.72 \\
\bottomrule
\end{tabular}
\caption{\label{tab-prompt_ablation}
VQA score of the answers generated by FLAN-T5-large conditioned on different prompts.
}
\end{table}
\end{document}